\documentclass{Interspeech}

\interspeechcameraready 

\title{Continual Speech Learning with Fused Speech Features}

\author[affiliation={1}]{Guitao}{Wang}
\author[affiliation={2}]{Jinming}{Zhao}
\author[affiliation={2}]{Hao}{Yang}
\author[affiliation={1}]{Guilin}{Qi}
\author[affiliation={2}]{Tongtong}{Wu*}
\author[affiliation={2}]{Gholamreza}{Haffari}

\affiliation{Southeast University}{Nanjing}{China}
\affiliation{Monash University}{Melbourne}{Australia}
\email{\{220222117, gqi\}@seu.edu.cn, \{firstname.lastname\}@monash.edu\\
 \quad \faEnvelope~*Corresponding author: tongtong.wu@monash.edu}


\usepackage{multirow}
\usepackage{booktabs}
\usepackage{float}
\usepackage{url}

\usepackage{cite}
\usepackage{hyperref} % 支持email链接
\usepackage{fontawesome}
\usepackage{xcolor}
\usepackage[normalem]{ulem}

\usepackage{comment}

\begin{document}

\maketitle

% the abstract here must exactly match the abstract entered into the paper submission system
\begin{abstract}
    % 1000 characters. ASCII characters only. No citations.
    Rapid growth in speech data demands adaptive models, as traditional static methods fail to keep pace with dynamic and diverse speech information.
    We introduce continuous speech learning, a new set-up targeting at bridging the adaptation gap in current speech models.
    We use the encoder-decoder Whisper model to standardize speech tasks into a generative format. We integrate a learnable gated-fusion layer on the top of the encoder to dynamically select task-specific features for downstream tasks.
    Our approach improves accuracy significantly over traditional methods in six speech processing tasks, demonstrating gains in adapting to new speech tasks without full retraining.
    % \footnote{\url{https://github.com/Kotiya-Qrank/ContinualSpeechLearning}}
    % Implementing continual learning in speech technologies enhances adaptability and efficiency, paving the way for advancements in real-world applications and further research in adaptive learning methods. 
\end{abstract}

\section{Introduction}

% 写法为由持续学习引入语音
% 那么就不要在一开始就提到语言，先简要减少完CL，再说CL在语音没有充分探索，提出CL在语音的挑战。
% 我认为没必要在这里介绍持续学习的三种方案 ×
% 是否有必要将method部分提到不同layer有不同信息写到introduction里面？ √

% 持续学习在NLP和CV获得巨大成功
% 持续学习的主要方法
% 然而，持续学习在语音上的应用很少
% 持续学习在语音上有独特的挑战
% 为此，我们提出方法...

Continual learning (CL)~\cite{abs-2402-01364,SustekSH22,WangZSZ24} has emerged as a promising solution to enable models to adapt to new data streams while retaining previously learned knowledge. While CL has been widely studied in natural language processing (NLP)~\cite{ke2022continual} and computer vision (CV)~\cite{qu2021recent}, its application to speech processing remains largely unexplored. Traditional speech models, trained on fixed datasets, often fail to generalize to new conditions, requiring costly full retraining. 
% \red{Compared to text, speech data is inherently more complex, giving rise to a wider variety of tasks. Adding new tasks is more challenging than adapting to new domains.}
This calls for a unified approach capable of handling multiple speech-related tasks by continual learning.

% CL has demonstrated significant success in fields like natural language processing (NLP)~\cite{ke2022continual} and computer vision (CV)~\cite{qu2021recent}. 
% Despite of this, its application to speech processing remains largely unexplored.

% 需要参考文献支撑，不同domain，类型。。。new data streams开展讲讲
% 这里写cl的survey
% \guitao{
Existing CL algorithms~\cite{abs-2402-01364,abs-2401-16386,abs-2302-00487} can be categorized into three main groups.
Replay-based methods~\cite{rolnick2019experience, lopez2017gradient, chaudhry2018efficient} store previous task samples for joint training;
Regularization-based methods~\cite{kirkpatrick2017overcoming, li2017learning} impose parameter change constraints; 
Architecture-based methods~\cite{mallya2018piggyback, serra2018overcoming, wang2022learning} allocate dedicated parameters per task. While effective in NLP and CV, these methods face unique challenges in speech processing due to the hierarchical nature of speech representations.

Whisper~\cite{radford2023robust}, a pre-trained encoder-decoder speech model, serves as a strong foundation for continual speech learning. However, its pre-training is optimized for generative tasks like automatic speech recognition (ASR) and speech translation (ST), which primarily capture content-related features. Prior studies~\cite{yang2023investigating} have shown that different layers of a speech model’s encoder encode distinct types of information, such as content, speaker identity, and paralinguistic cues. Notably, higher layers focus on content, while lower and intermediate layers retain speaker-related and prosodic features. This layer-wise specialization poses a critical challenge for continual learning~\cite{WuCLLQH22}, as naive fine-tuning risks overwriting speaker and paralinguistic signals in intermediate layers, leading to information loss.

In this paper, we introduce continual speech learning (CSL), a new setup designed to handle the multifaceted nature of speech, encompassing content, speaker characteristics, paralinguistic features, and semantics. 
To achieve this, we integrate a learnable Gated-Fusion Layer (GFL) on top of Whisper's encoder
% 相似的任务共享参数迁移，不同的任务可以做到参数隔离
enabling dynamic fusion of diverse speech features for individual tasks. This mechanism facilitates parameter sharing across related tasks while preserving task-specific representations, ensuring better generalization and adaptability. By standardizing speech tasks into a unified generative format and leveraging GFL, our approach significantly enhances the model’s ability to distinguish between diverse speech tasks while effectively adapting to varying speech characteristics. 
This results in improved performance across tasks without compromising previously learned knowledge, making GFL a robust solution for continual speech learning.
% 
% By standardizing speech tasks into a unified generative format and incorporating the GFL, we significantly improve the model's ability to distinguish between diverse speech tasks while effectively adapting to varying speech characteristics.

% We conduct extensive study on applying CL methods popular in the NLP field on a wide range of speech tasks. To the best of our knowledge, this is the first attempt in speech. 
% % 
% We propose integrating WSL in speech encoding, which fuses speech features and improves the model's ability to differentiate between diverse speech tasks, outperforming traditional continual learning methods.
% % 
% We analyze key factors influencing the continual learning performance, shedding lights on how speech models can be better adapted for dynamic environments.
Our key contributions are:
(i) The first comprehensive study on applying CL methods to diverse speech tasks, bridging the gap between cross-modal CL and speech processing.
(ii) A novel framework integrating GFL into speech encoding, enabling dynamic fusion of heterogeneous speech features for superior task differentiation and outperforming conventional CL baselines.
(iii) A rigorous analysis identifying key factors affecting CL performance in speech, providing guidelines for adapting speech models to real-world dynamic environments.

% Our key contributions are summarized as follows:
% % 
% (i) We conduct the first comprehensive study on applying CL methods to diverse speech tasks, bridging the methodological divide between cross-modal CL and speech processing.
% % 
% (ii) We propose a framework that incorporates GFL into speech encoding. This approach dynamically fuses heterogeneous speech features, significantly improving task differentiation and outperforming conventional CL baselines. 
% % 
% (iii) Through rigorous analysis, we identify key factors influencing continual learning performance in speech, offering guidelines for adapting speech models to dynamic real-world environments.

% \mich{where is your related work? You should also explain more on catastrophic forgetting, and explain a bit on how our method can mitigate the issue.}
% \guitao{i put related works in paragraph 1}

\section{Continual Speech Learning Setup}

% \mich{guitao: why didn't you follow the structure of this paper referred by Tong: https://arxiv.org/pdf/1706.08840. Please try imitating their way of writing (i.e., changed II and III) unless you have a good reason for not doing so.}
% \guitao
% {
% % 第二部分一般称为Preliminaries，内部分为两个子章节，任务定义，持续学习定义，第三部分介绍自己的方法。
% % https://doi.org/10.1609/aaai.v37i11.26492
% % https://proceedings.neurips.cc/paper_files/paper/2023/file/398b00a05b847ac65eb98c8e5e865fe8-Paper-Conference.pdf
% % 我不太清楚我们的任务定义应该怎么写，因为我们融合了多个任务，还是说写语音信息抽取的定义，正如intro里提到的。
% % 有的时候第二部分也会介绍metrics，但我们评价指标有区别，不能完全套用。正如你提到的这篇https://arxiv.org/pdf/1706.08840
% % 这篇文章也是多任务，写法和我比较类似，先介绍持续学习定义，然后指出和一般持续学习的不同。https://arxiv.org/pdf/2401.08295
% mich:
% The second part is generally referred to as Preliminaries, which is divided into two sub chapters: task definition and Problem Formulation(continual learning). The third part introduces one's own methods.
% https://arxiv.org/pdf/2211.11226
% https://arxiv.org/pdf/2310.04801
% I'm not quite sure how to write our task definition, as we have integrated multiple tasks, or definition for speech information extraction, as mentioned in the intro.
% Sometimes the second part also introduces metrics, but our evaluation metrics differ and cannot be fully applied. As you mentioned above https://arxiv.org/pdf/1706.08840.
% This article is also multitasking, with a writing style similar to mine. Firstly, it introduces the definition of continuous learning, and then points out its differences from general continuous learning. https://arxiv.org/pdf/2401.08295.
% }

Continual learning sequentially trains a model on a series of tasks. 
Given a sequence of task \( \{\mathcal{T}_1, \mathcal{T}_2, \ldots, \mathcal{T}_N\} \), where \( N \) is the number of tasks.
% each task contains a separate dataset and has its own training set, validation set, and test set. 
Each task has its own dataset, including training, validation, and test sets.
% \red{All datasets are mutually exclusive across tasks.}
At the $k_{th}$ step, the model is trained on $\mathcal{T}_k$ to learn new knowledge and will be evaluated on $\mathcal{\widetilde{T}}^{test}_k = \bigcup_{i=1}^k \mathcal{T}^{test}_i$.
% The major challenge for continual learning is catastrophic forgetting, where model performances bad on old tasks when trained on new ones. 
Unlike traditional continual learning in NLP and CV, where tasks are typically created by partitioning a single dataset into different subsets~\cite{wang2024comprehensive}, 
we use entirely distinct datasets to define each task because speech data is more complex, leading to a wider variety of tasks\cite{yang2021superb, WuWZLQLH22, kang2024event}. Adding new tasks is more challenging than just adapting to new domains. In real-world, domain shifts naturally extend this challenge, requiring methods that can adapt to evolving tasks.

% 暂时隐藏
% \begin{table*}[t]
% \caption{Dataset Statistics}
%     \centering
%     \resizebox{.9 \textwidth}{!}{
%         \begin{tabular}{c|c|c|c|c}
%             \toprule
%             \multirow{2}{*}{ID} & \multirow{2}{*}{Name} & \multirow{2}{*}{Category} & \multirow{2}{*}{Dataset} & Instances \\
%              & & & & ( Train \textbar\textbar\space Dev \textbar\textbar\space Test ) \\
%             \midrule
%             KS  & Keyword Spotting  & Content & Speech Commands & 22246 \textbar\textbar\space 3093 \textbar\textbar\space 3081 \\ 
%             SID & Speaker Identification & Speaker & VoxCeleb1\_top10 & 5379 \textbar\textbar\space 51 \textbar\textbar\space 56 \\ 
%             ER  & Emotion Recognition & Paralinguistics & IEMOCAP & 3555 \textbar\textbar\space 890 \textbar\textbar\space 1085 \\ 
%             IC  & Spoken Intent Classification & Semantics & Fluent Speech Commands &  23132 \textbar\textbar\space 3118 \textbar\textbar\space 3793 \\ 
%             SF  & Spoken Slot Filling & Semantics & SNIPS & 104672 \textbar\textbar\space 2800 \textbar\textbar\space 2800 \\ 
%             ASR & Automatic Speech Recognition & Recognition & LibriSpeech & 28539 \textbar\textbar\space 2703 \textbar\textbar\space 2620 \\ 
%             \bottomrule
%         \end{tabular}
%     }
%     \label{tab:dataset}
% \end{table*}

\section{Methodology}

\subsection{Model Architecture and Modification}
% 
% We use encoder-decoder Whisper-base model as our backbone model. The pre-training tasks of Whisper include transcribe and translate. Transcribe involves recognizing English speech, while translate involves translating speech from other languages into English. For generating, Whisper needs to specify the task, together with language and timestamps. Whisper generates based on these prefix tokens. 
% For our tasks, we made the following modifications:

We adopt \textsc{Whisper-base}\footnote{https://huggingface.co/openai/whisper-base} with encoder-decoder architecture as our backbone model, which is a multitask model primarily pre-trained on transcribing and translating tasks. Whisper executes tasks by using the corresponding prefixes (e.g.,``transcribe" and ``translate") as prompts to guide the model. 
% \mich{(need to cite papers that prompt Whisper. Check Hung-yi Lee and Shinji Watanabe's papers)}.\guitao{i dont understand here, what i mentioned above is basic whisper} \mich{some other papers also study prompts for Whisper. We should cite them.}
% \hao{introduce whisper, no more details on translation task. and do not inculde many details, try to answer: what's the pre-training protocol? how to handle multitask?} 
% Whisper’s pretraining tasks include transcribing and translating. 
% Transcribing involves recognizing English speech, while translating entails converting speech from other languages into English. 
% Whisper requires task specifications, including language and timestamps, to perform these tasks, generating output based on prefix tokens.
% \guitao{Whisper is a multitask model pre-trained on transcribing and translating tasks. Whisper manages multitasking by using task-specific prefixes that guide it in differentiating between tasks, including language, task type and whether to use timestamps.}

For our continual learning setting, we make the following modifications:
\emph{(1) Task-specific tags.} To recognize and decode the current task, we introduce additional task-specific tags into Whisper's vocabulary.
% in addition to its predefined tags. 
For example, we use \textless\textbar KS\textbar\textgreater\space for the Keyword Spotting task.
\emph{(2) Special tokens}. We also introduce special tokens which are needed to label speakers in Speaker Identification, and slot categories in Slot Filling, respectively.
% 
% For our continual learning tasks, we add Task-specific tags like "\textless\textbar IC\textbar\textgreater" and special tokens for SID and SF.
% 暂时隐藏
% \noindent\textbf{Task-specific tags:}
% We focus on continual learning across multiple tasks, where speech inputs might be similar, but the desired outputs differ depending on the task. \hao{remove the above, explain: to make model recognize different tasks, so introduce different tags.} To help the model distinguish between tasks, we introduce additional task-specific tags beyond the predefined "transcribe" and "translate" tasks in Whisper. For example, we add tags like "\textless\textbar KS\textbar\textgreater", "\textless\textbar IC\textbar\textgreater", "\textless\textbar ER\textbar\textgreater", etc. and include these in Whisper’s vocabulary \hao{to replace \textless transcribe \textgreater ?}.
% % 暂时隐藏
% \noindent\textbf{Special tokens:} \hao{remove this paragraph, move this to last paragraph, explain that why you add tags to vocab, and why you add special tokens to vocab.}
% This modification is mainly for the SID and SF tasks. As mentioned earlier, SID introduces speaker tags for different speakers, while SF adds tokens representing various slot categories. To handle these tasks, we expand the model’s vocabulary to include these additional special tokens.

\begin{figure}[t]
    \centering
    \includegraphics[width=\linewidth]{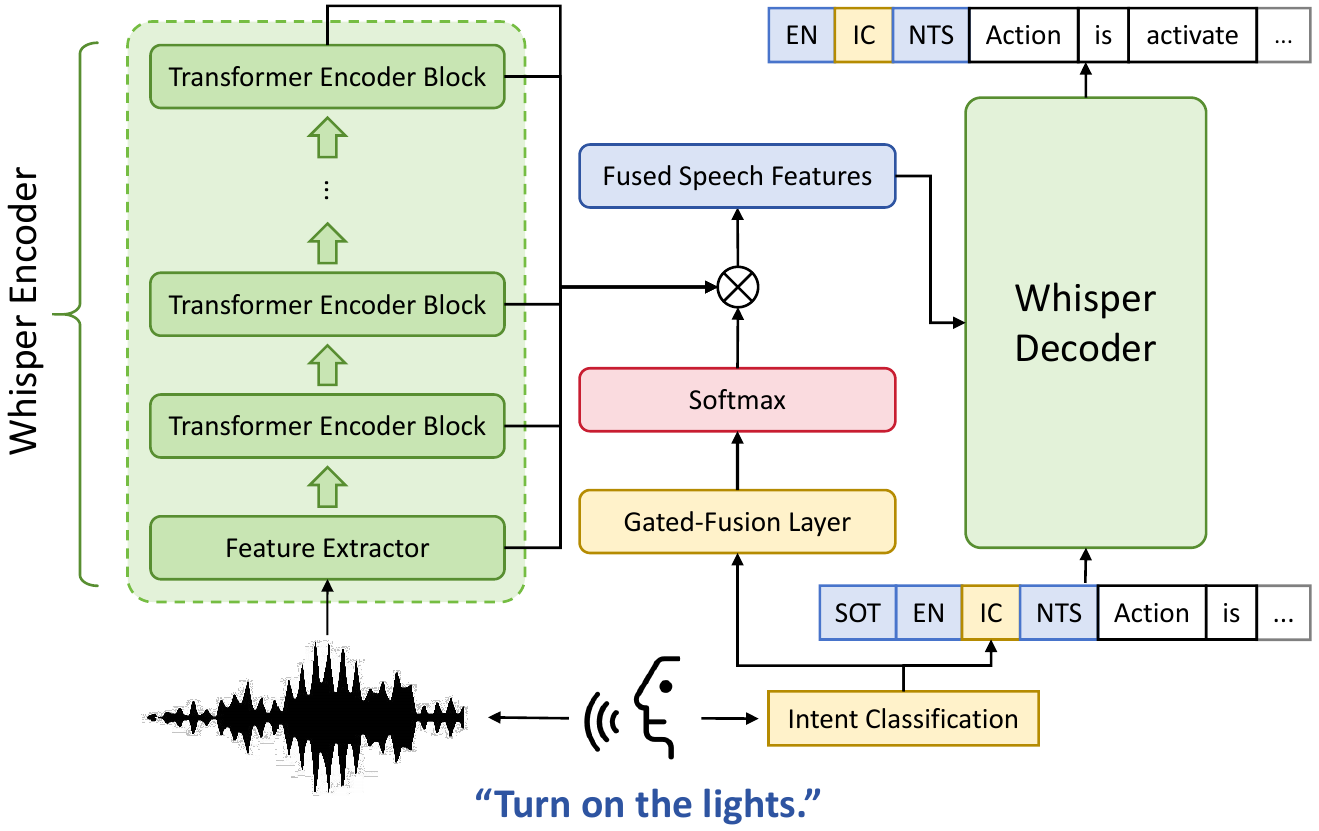}
    \caption{Framework of Continual Speech Learning with Fused Speech Features. 
    % \mich{this figure takes up too much space. Please make it smaller. Please add more descriptions (highlight key points).}
    }
    \label{fig:framework}
\end{figure}

\subsection{Fusion of Speech Features}
Previous studies~\cite{yang2023investigating} have shown that different layers of Whisper's encoder store distinct types of information. 
To meet the need for various types of information under a continual learning setting, 
% To leverage this in continual learning,
we propose \emph{Continual Speech Learning with Fused Speech Features}.
% \footnote{\url{https://github.com/Kotiya-Qrank/ContinualSpeechLearning}}
\footnote{https://github.com/Kotiya-Qrank/ContinualSpeechLearning}

We incorporate a Gated-Fusion Layer (GFL) into Whisper's encoder to fuse multi-layer features as speech representations, enabling the model to effectively capture and utilize diverse speech features for various speech tasks. 
Figure \ref{fig:framework} illustrates our framework.
% 
% This approach computes a weighted sum of the hidden states from encoder layers as speech representations for decoding.
% feeds the result to the decoder.
% 
Specifically, we concatenate the normalized outputs from each layer of Whisper's encoder to form a unified representation of the hidden states, denoted as $\mathbf{H}^{norm}$:
% \hao{remove the process about norm, explain the computing process(you create a trainable linear layer, every task has one weighted-sum layer, or a shared weighted-sum layer?), and put , after Equation.} 
% The process begins by normalizing the output from each layer of Whisper's encoder and concatenating them to form a unified representation of the hidden states, denoted as \( H^{norm} \).
\begin{equation}
    \boldsymbol{h}^{norm}_{i} = \mathrm{LayerNorm}(\boldsymbol{h}_{i}),
\end{equation}
\begin{equation}
    \mathbf{H}^{norm} = [ \boldsymbol{h}^{norm}_{1}, \boldsymbol{h}^{norm}_{2},...,\boldsymbol{h}^{norm}_{M} ],
\end{equation}
where \( \boldsymbol{h}_{i} \) is the hidden state from the \( i \)-th layer. 
% \( d \) is the dimension of the hidden layer. \( H \in R^{n \times d} \) is the collection of all the representations, 
\( M \) is the number of layers of Whisper's encoder.
 % \mich{(Notation is unclear. Does h\_i refer to representation for a single token x, or an entire input sequence \textbf{x}? Please rewrite this part )}
% For each layer's weights, we also normalize them so that their probabilities sum up to one. Then, we multiply them by the hidden states to obtain the final weighted sum of the hidden states.
% Then, we build a trainable linear layer \( \mathbf{W} \in R^{n \times T} \) \mich{(you don't seem to mention what is t)} to fuse the different hidden states.
% 
% \( T \) represent the total number of tasks. 
For task \( k \), \( \boldsymbol{t}_k \in \mathbb{R}^{1 \times N}\) is the one-hot vector representing the task.
We use temperature \( t \) to perform softmax on $\mathbf{W}\in \mathbb{R}^{N \times M}$, the parameter matrix of our proposed GFL method. Temperature 
\( t \) controls the smoothness of the softmax output. We then calculate the final speech representation \( \boldsymbol{h}_k \): 
% \mich{( 1) I don't understand what ${t_j}$ represents. 2) Shouldn't you include temperature in the formula? 3) Your notation is inconsistent. In Section 2, the task is notated as ${T_k}$, but here it's j-th task. Please make them consistent. 4) w should be bolded. When it's not bolded, it represents scalar. )}. 
% Then, we calculate the weighted sum of all layers to obtain the final speech representation \( h \) for given task.
\begin{equation}
    \boldsymbol{h}_k = \boldsymbol{t}_k \cdot \mathrm{SoftMax}(\mathbf{W}, t) \cdot \mathbf{H}^{norm},
    % \quad \text{where } \bm{t}_k = 
    % \begin{cases} 
    % 1 & \text{if } k = current \ task, \\
    % 0 & \text{otherwise}.
    % \end{cases}
\end{equation}
% 
% \begin{equation}
%     \bm{w}_k = \bm{W}^\top \bm{e}_k, \quad \text{where } (\bm{e}_k)_i = 
%     \begin{cases} 
%     1 & \text{if } i = k, \\
%     0 & \text{otherwise}.
%     \end{cases}
% \end{equation}
% 
% 
% where \( w \in R^{n} \) is the weight of encoder layers and \( h \) is the final representation passed to the decoder.
% This approach computes a weighted sum of the hidden states from encoder layers as speech representations for decoding.
% 
We combine the GFL with experience replay method to review past knowledge during training, enabling the decoder to adapt to features from multiple tasks simultaneously.
% \mich{(please briefly explain what it means, and why it's helpful. Also use citations here)} 
% enhancing CSL by efficiently utilizing features from multiple encoder layers.

We propose two different training strategies for GFL:
 (1) {Single} stage learning (i.e., GFL$_{S}$) initialize the parameters of GFL with zero, then jointly train the parameters of the GFL and the decoder, balancing their learning via a temperature parameter.
 (2) {Double} stage consolidation (i.e., GFL$_{D}$) initialize a new model with the GFL parameters trained in GFL$_{S}$, then freeze the GFL and trains only the decoder.

% \guitao{
% Our method has two training strategies.
%  (1) \emph{WSL}: Jointly train the parameters of the WSL and the decoder, balancing their learning via a temperature parameter.
%  (2) \emph{Pre-defined WSL}: Initialize a new model with the WSL parameters trained in the first approach, then fixes the WSL and trains only the decoder.
% }

% As can be seen in Table ~\ref{tab:output}.

% \begin{table*}[t]
%     \centering
%     \resizebox{.9 \textwidth}{!}{
%         \begin{tabular}{c|c|c|p{8cm}}
%             \toprule
%             ID & Dataset & Category & Output Example \\
%             \midrule
%             KS  & Speech Commands        & discriminative & \textbf {go} \\ 
%             SID & VoxCeleb1\_top10       & discriminative & Speaker is \textbf {\textless\textbar speaker934\textbar\textgreater}. \\ 
%             ER  & IEMOCAP                & discriminative & Emotion is \textbf {neutral}.   \\ 
%             IC  & Fluent Speech Commands & discriminative & Action is \textbf {change language}. Object is \textbf {none}. Location is \textbf {none}.     \\ 
%             SF  & SNIPS                  & generative     & \textbf {Book a B-restaurant\_type food truck E-restaurant\_type in B-city argusville E-city that has B-served\_dish fish chips E-served\_dish}. \\ 
%             ASR & LibriSpeech            & generative     & \textbf {Public prosecutor and determined jury sat every day their lists went forth every evening}. \\ 
%             \bottomrule
%         \end{tabular}
%     }
%     \caption{Dataset Output Example}
%     \label{tab:output}
% \end{table*}

\begin{table*}[t]
    \centering
    \caption{
% Main results of each task after training on the last task in terms of their corresponding metrics. \hao{fr is the percentage of decrease, or the number decrease? FT: 0.00, 52.91, it looks weird.}\guitao{FR means the number decrease from the top score during the training.}
Performance comparison of various methods on different tasks.
Accuracy (Acc), Slot Type F1-score (STF1), and Word Error Rate (WER) are reported after training on the last task. 
% Acc represents accuracy, STF1 represents Slot Type F1-score, WER represents Word Error Rate, and 
Average Forgetting (AF) denotes the average drop in performance from the highest score during training. 
% 
% To provide an overall assessment, we compute the Mean Rank (MR) for each method.
Mean Rank (MR) represents the average performance rank across all tasks.
% \mich{(no need to explain abbreviations if they are mentioned in text already)}. 
% S means single stage learning, D means double stage consolidation . 
} 
    \vspace{-.1cm}
    \resizebox{.85\textwidth}{!}{
        \begin{tabular}{c|cc|cc|cc|cc|cc|cc|c}
            \toprule
            \multirow{2}{*}{Method} & \multicolumn{2}{c|}{KS}& \multicolumn{2}{c|}{SID} & \multicolumn{2}{c|}{ER} & \multicolumn{2}{c|}{IC} & \multicolumn{2}{c|}{SF} & \multicolumn{2}{c|}{ASR} & Overall \\
             & Acc$\uparrow$ & AF$\downarrow$ & Acc$\uparrow$ & AF$\downarrow$ & Acc$\uparrow$ & AF$\downarrow$ & Acc$\uparrow$ & AF$\downarrow$ & STF1$\uparrow$ & AF$\downarrow$ & WER$\downarrow$ & AF$\downarrow$ & MR$\downarrow$ \\
            \midrule
            
            % IFT & 97.11 & - & 57.14 & - & 47.00 & - & \textbf{99.31} & - & \textbf{79.26} & - & \textbf{94.47} & - \\ 

            % WSL & \textbf{97.40} & - & \textbf{64.29} & - & \textbf{49.12} & - & 98.08 & - & 69.89 & - & 86.18 & - \\ 
            
            % \midrule
            
            % FT & 74.23 & 23.17 & 14.29 & 33.92 & 24.52 & 20.83 & 0.00 & 52.91 & 0.00 & 42.84 & 7.52 & - \\ 
            % MTL & 68.97 & - & 42.86 & - & 41.20 & - & 73.45 & - & 38.96 & - & 6.03 & - \\
            % ER & \textbf{95.20} & \textbf{1.62} & 32.14 & 21.43 & 32.81 & \textbf{7.10} & 7.91 & 47.46 & 10.30 & \textbf{9.26} & 7.87 & - \\ 
            % LwF & 81.82 & 15.58 & 12.50 & 33.93 & 41.29 & 13.00 & 0.00 & 99.08 & 2.70 & 73.30 & 6.18 & - \\
            % DER++ & 86.63 & 10.77 & 0.00 & 45.90 & 33.73 & 12.17 & 45.11 & 36.22 & 0.00 & 73.47 & \textbf{5.50} & - \\ 
            % Ours & 75.82 & 21.36 & \textbf{50.00} & \textbf{14.29} & \textbf{42.58} & 9.68 & \textbf{78.22} & \textbf{8.55} & \textbf{41.40} & 18.75 & 16.87 & - \\

            MTL & 89.42 & - & 58.93 & - & 63.32 & - & \textbf{99.55} & - & 78.39 & - & \textbf{5.78} & - & 3.34 \\
            FT & 81.79 & 15.16 & 16.07 & 25.00 & 52.35 & 16.68 & 33.72 & 61.48 & 0.00 & 75.11 & 7.60 & - & 6.00 \\ 
            Replay & \textbf{96.79} & \textbf{0.16} & 67.86 & 0.00 & 66.73 & 1.20 & 90.98 & 6.36 & 81.65 & 1.59 & 6.94 & - & 2.83 \\ 
            LwF & 60.18 & 37.22 & 35.71 & 39.29 & 49.31 & 17.33 & 32.38 & 45.71 & 0.00 & 84.50 & 8.49 & - & 6.67 \\
            DERPP & 95.75 & 1.20 & 73.21 & 1.79 & 68.29 & 0.56 & 98.44 & 0.27 & \textbf{82.66} & \textbf{1.06} & 10.30 & - & 3.00 \\ 

            \midrule
            
            GFL$_{S}$ & 82.38 & 15.12 & 82.14 & \textbf{-1.78} & 66.45 & 2.58 & 95.44 & \textbf{-10.44} & 79.59 & 3.54 & 7.72 & - & 3.83 \\
            GFL$_{D}$ & 95.26 & 1.88 & \textbf{83.93} & 8.93 & \textbf{68.39} & \textbf{-0.56} & 98.68 & 0.79 & 73.88 & 9.36 & 7.34 & - & \textbf{2.50} \\

% MTL   4+5+5+1+4+1=20
% FT    6+7+6+6+7+4=36
% ER    1+4+3+5+2+2=17
% LWF   7+6+7+7+7+6=40
% DERPP 2+3+2+3+1+7=18
% S     5+2+4+4+3+5=23
% D     3+1+1+2+5+3=15

            \bottomrule
        \end{tabular}
}

\label{tab:results}
\end{table*}

\section{Experiment}

\subsection{Dataset}

% We conducted experiments on various tasks from SUPERB, each task has its unique dataset, including Speech Commands from KS, VoxCeleb1\_top10 from SID,  IEMOCAP from ER,  Fluent Speech Commands from IC, SNIPS from SF and LibriSpeech from ASR. 

% \subsection{Dataset Construction}
% 
We conduct our experiments on six tasks with the corresponding datasets from SUPERB~\cite{yang2021superb}. 
% We show our dataset statistics in table \ref{tab:dataset}.
% We selected a subset \hao{(explain what subset/subsets)} of dataset from SUPERB Benchmark for our Continual Speech Learning experiments.
% \hao{(move this to the beginning of this paragraph)}
% 
These tasks can be categorized into discriminative (e.g., IC) and generative tasks (e.g., ASR). To align the decoding strategies for these two types of tasks, we decode discriminative tasks in the same way of generative tasks. 
These tasks are:

\noindent{\bf Keyword Spotting (KS)}
% detects preregistered keywords by classifying utterances into a predefined set of words. 
We used Speech Commands dataset v1.0~\cite{warden2017speech} which consists of ten classes of keywords, a class for silence, and an unknown class to account for false positives.
% The task output is the detected command directly.

\noindent{\bf Speaker Identification (SID)} 
% classifies each utterance based on its speaker identity using a multi-class classification approach. 
The popular VoxCeleb1~\cite{nagrani2020voxceleb} dataset is used. 
% Noticed that Whisper performs poorly on this task. We decided to use the top ten speakers with the most speech data as a new dataset named VoxCeleb1\_top10.
% Original VoxCeleb1 dataset has over a thousand categories, where the speech is a sentence, and the output is the corresponding person's name.
% Since different names don't have any special meaning, we sequentially assigned a number to each person and output a special identifier like \textless\textbar speaker1\textbar\textgreater\space and \textless\textbar speaker2\textbar\textgreater.
% But given that Whisper performs poorly on this task, we created a new dataset, VoxCeleb1\_top10, which includes the top ten speakers with the most speech data.
% The original VoxCeleb1 dataset contains thousands of categories, which is difficult for Whisper. We use the top ten speakers as a new dataset named VoxCeleb1\_top10. 
% Since different names don't have any special meaning, for our setup, we assigned a number to each speaker and output a special identifier like \textless\textbar speaker1\textbar\textgreater.
We use the top ten speakers as a subset named VoxCeleb1\_top10.
% because Whisper performs poorly on the origin thousands of categories.
We assign a unique label to each speaker, such as \textless\textbar speaker1\textbar\textgreater.

% \verb|<|speaker1\verb|>|
% \verb|<|speaker1\verb|>|
% \textless\textbar speaker1\textbar\textgreater

\noindent{\bf Emotion Recognition (ER)} 
% predicts an emotion class for each utterance.
We use IEMOCAP~\cite{busso2008iemocap}, which contains four emotional classes, including ``happy", ``sad", ``neutral" and ``angry".
% \mich{(what are these classes?)}
% task where the speech is a sentence, and the output is the corresponding emotion. 
% We add the prefix token ”Emotion is” before the emotion token.

\noindent{\bf Intent Classification (IC)}
% 
% IC categorizes utterances into predefined classes to determine the speakers’ intent. The dataset is Fluent Speech Commands, where each utterance is tagged with three intent labels: action, object, and location.
% 
% Fluent Speech Commands consist of three multi-classification tasks, where all three categories need to be correct together. We choose to add corresponding prefix tokens for each category, such as ”Action is xxx.”. Therefore, the output is a combination of three sentences with prefix tokens and category tokens
% 
% categorizes utterances into predefined classes to determine the speakers' intent. 
The dataset is Fluent Speech Commands~\cite{lugosch2019speech}, where each utterance is tagged with three intent labels: action, object, and location.
% All three categories need to be correct together. 

\noindent{\bf Slot Filling (SF)}
% 
% SF predicts a sequence of semantic slot-types from an utterance, like a slot-type FromLocation for a spoken word Taipei, which is known as a slot-value. The dataset used is Audio SNIPS dataset, which synthesizes multi-speaker utterances for SNIPS.
% 
% SF requires to label each word while translating, we follow the SUPERB’s approach and use special identifiers like ”B-city” and ”E-city” to enclose corresponding words.
% 
% predicts a sequence of semantic slot-types from an utterance, like a slot-type ``FromLocation'' for a spoken word Taipei, which is known as a slot-value. 
We use the Audio SNIPS~\cite{lai2021semi}, which synthesizes multi-speaker utterances.
We use special tokens, such as ``B-city" and ``E-city", to mark slot-types in the utterance.

\noindent{\bf Automatic Speech Recognition (ASR)}
% transcribes speech into text. 
We use a subset of LibriSpeech~\cite{panayotov2015librispeech} dataset. While ASR is Whisper's pre-training task, we re-format samples in the dataset by adjusting case and adding punctuation.
% , as the original dataset is in all uppercase letters.

% The randomly selected task sequence for CSL is: KS → SID → ER → IC → SF → ASR. 

% This progression starts with simpler recognition tasks and gradually moves towards more complex semantic and generative tasks.

\subsection{Baselines}

We compare our method with several baselines which are commonly used in continual learning: 
% \hao{it's better to differentiate cl method training and without cl method training}
% 
% (1) \emph{Independent Fine-tuning (IFT)}: Fine-tune the Whisper model on each task independently without considering other tasks or replaying previous ones.
% 
% (2) \emph{Weighted Sum Layer Fine-tuning (WSL)}: Fine-tune the Whisper model on each task independently using the WSL to fuse speech features from multiple encoder layers. \mich{please rename this setting}
% 
(1) \emph{Multi-task Learning (MTL)}: Jointly fine-tune the Whisper model on all tasks simultaneously.
(2) \emph{Fine-tuning (FT)}: Sequentially fine-tune the Whisper model on each task without replaying previous tasks.
(3) \emph{Experience Replay (Replay)~\cite{rolnick2019experience}}: A memory buffer stores and replays selected examples from previous tasks during training on new tasks.
(4) \emph{Learning without Forgetting (LwF)~\cite{li2017learning}}: Retain knowledge from previous tasks by keeping an old model and computing a knowledge retention loss.
(5) \emph{Dark Experience Replay ++ (DERPP)~\cite{buzzega2020dark}}: Store logits of previous task outputs and use knowledge distillation to replay them and compute the loss for previous tasks.

\subsection{Evaluation Metrics}

% We use accuracy (Acc) for all discriminative tasks, slot type F1 score (STF1) for SF and word recognition rate (WRR) for ASR at the end of the task sequence.
% Considering continual learning, we compute Forgetting Rate (FR) for each task.

We use the following metrics to evaluate the performance of our model across tasks: (1) \emph{Metrics for assessing task performance}~\cite{yang2021superb}: 
Accuracy (Acc) for all discriminative tasks.
Slot Type F1 score (STF1) for SF.
% Word Recognition Rate (WRR) for ASR. 
Word Error Rate (WER) for ASR. 
% WRR is the complementary metric of Word Error Rate (WER), defined as WRR=1-WER, which directly reflects the proportion of correctly recognized words. A higher WRR indicates better system performance.\hao{explain why not WER}
% 
(2) \emph{Metric for assessing learning process}:  
% Forgetting Rate (FR)~\cite{chaudhry2018riemannian} measures the difference between the maximum knowledge gained about the task throughout the learning process in the past and the knowledge the model currently has about it. 
Average Forgetting (AF)~\cite{chaudhry2018riemannian} measures the difference between the maximum past performance and the current performance on a particular task.
Let $a_{k,j}$ be the testing performance on the $j\text{-}th$ task after training from tasks $1$ to $k$, forgetting for the $j\text{-}th$ task is computed with
$f_j^k = \max_{l \in \{1, \dots, k-1\}} a_{l,j} - a_{k,j}, \quad \forall j < k$. 
% The overall forgetting is the average of $f_j^k$ across all tasks at the end of training.
The average forgetting at $j\text{-}th$ task is written as
$AF_k = \frac{1}{k - 1} \sum_{j=1}^{k - 1} f_k^j$.
(3) \emph{Metric for overall assessment}:  
Mean Rank (MR) calculates the average rank of each method across all tasks.  
% Mean Rank
% To holistically assess model performance, we introduced the Mean Rank (MR) metric, which calculates the average rank of each method across all tasks.
Let $\text{Rank}_{i,j}$ be the testing performance rank (e.g., 1, 2, 3) of method $i$ on the $j$-th task.
The Mean Rank of method $i$ is computed as:
$\text{MR}_i = \frac{1}{N} \sum_{j=1}^{N} \text{Rank}_{i,j}$,
where $N$ is the total number of tasks.

% \begin{equation}
%     FR_j^k = f_j^k = \max_{l \in \{1, \dots, k-1\}} a_{l,j} - a_{k,j}, \quad \forall j < k.
% \end{equation}

% , indicating how much the model's performance deteriorates on previous tasks after learning new ones.

% \begin{equation}
%     Forgetting Rate = 
% \end{equation}

% 这是GPT告诉我的
% Word Recognition Rate (WRR)
% 公式: 
% WRR=1−WER
% WRR 是 WER 的互补指标，表示正确识别的单词比例。它直接反映了准确率，WRR 越高，表示系统的性能越好。
% 例如，如果 WER = 0.2，WRR = 1 - 0.2 = 0.8 (即 80%)。

\subsection{Implementation Details}
% \mich{(this section can be shortened if we run out of space.)}
% We use pretrained \textsc{Whisper-base} as the backbone model to carry out our experiments.
% We train the models 40 epochs for SID, 60 epochs for ER, and 20 epochs for other tasks. 
% The batch size is set to 16. 
% We use AdamW with an initial learning rate of 1e-6, with warmup rate at 10\%.
% We freeze Whisper’s encoder and the trainable parameters only includes Whisper's decoder and Weighted Sum Layer in our method.
% The maximum memory buffer for all tasks is set to 1000.

% We use the pre-trained Whisper-base as the backbone model for our experiments. 
We use different maximum epochs for each task due to limited training data: 40 for SID, 60 for ER, and 20 for others. 
The patience for early stopping is 1000 steps.
% \hao{why differnet epoch?}
% We set batch size to 16 and use the AdamW~\cite{loshchilov2017decoupled} optimizer with an initial learning rate of 1e-4, applying a learning rate warmup over the first 10\% of training steps. 
% The batch size is 16 and use the AdamW~\cite{loshchilov2017decoupled} optimizer with a 1e-4 learning rate, applying a warmup for the first 10\% of training steps.
The batch size is 16, and we use the AdamW~\cite{loshchilov2017decoupled} optimizer with a 1e-4 learning rate and 10\% warmup.
Whisper’s encoder is frozen during training.
The temperature of softmax is set to 0.0005.
The maximum memory buffer size of experience replay methods is 1000 instances.
% Differ from classification tasks, generative tasks requires more space to store logits. For DERPP method, the space needed is already much larger than the size of model. Therefore, we store old models instead of logits when applying DERPP.
% Our approach has two different training strategies.
%  (1) WSL$_{S}$: The parameters of WSL is initialized with zero, then jointly train the parameters of the WSL and the decoder, balancing their learning via a temperature parameter.
%  (2) WSL$_{D}$: Initialize a new model with the WSL parameters trained in the first approach, then frozen the WSL and trains only the decoder.
We prepend prompts to the output results and apply constraint decoding during inference (detailed in Section \ref{sec:Prompts}.)

\section{Results and Analysis}

\subsection{Analysis of Gated-Fusion Layer }

% 大多数方法在KS和ASR这两个平均效果很好的任务上面依旧表现很好，然而在流程中间的任务上效果很差，甚至有些方法完全遗忘了某些任务。与传统的方法恰恰相反的是，我们的方法虽然在KS和ASR上表现平庸，但保持了在每一个任务上的高水准，尤其在IC和SF上最为明显，远远优于其他方法。这证明了我们的方法可以兼顾到持续学习中的每一个任务，不至于太受任务形式、数据集大小所影响。

% 先介绍baseline的整体情况，
% 直接ft方法的效果很差，说明csl的灾难性遗忘非常严重，
% mtl的结果很在每个任务上都很稳定，但表现都不是很突出，除了ASR
% ER和DERPP的整体表现都很好，ER在SID任务上的效果较差，而DERPP在ASR上效果较差，

We compare results of various continual learning methods after training on the last task in Table~\ref{tab:results}. Our proposed GFL framework demonstrates strong continual learning capabilities by dynamically selecting task-specific features while mitigating catastrophic forgetting. Compared to baseline methods, GFL$_{D}$ achieves the highest accuracy across multiple tasks (e.g., 83.93\% in SID, 68.39\% in ER). 
% and exhibits negative forgetting in these tasks, indicating an ability to improve rather than simply retain knowledge.
GFL$_{S}$ also performs well, effectively balancing forgetting reduction with competitive accuracy. 
GFL methods show negative AF scores in some tasks, indicating performance improvement on prior tasks through positive knowledge transfer.
In contrast, fine-tuning suffers from severe forgetting, while Replay and DERPP mitigate forgetting but do not achieve the same accuracy as GFL$_{D}$. LwF struggles with knowledge retention, showing high AF values across tasks. Although MTL achieves high performance on certain tasks (e.g., 99.55\% in IC, 5.78 WER in ASR), it lacks practical applicability in continual learning, as it requires simultaneous training on all tasks. Furthermore, MTL often faces gradient conflicts where optimizing one task can harm another.

% The direct FT method performs poorly, indicating that catastrophic forgetting in CSL is particularly severe.
% The results of MTL are stable across all tasks, but not outstanding, except for ASR. In multi-task learning, the model is required to handle multiple tasks simultaneously. However, the speech tasks exhibit significant differences, which may lead to task interference, negatively impacting the performance of each task.
% Both ER and DER++ perform well overall, with ER showing relatively poor performance on the SID task, while DER++ performs poorly on the ASR task.

% The most promising results come from the two variants of the WSL approach. GFL$_{S}$ achieves a balance between accuracy and forgetting. On SID and IC, the results are not only not forgotten, but even improved.
% The GFL$_{D}$ method yields even higher accuracy and lower forget rate on most tasks, and performs well on the other tasks, indicating that leveraging pretrained WSL can significantly enhance performance and reduce forgetting. 

Most methods achieve strong performance on tasks with inherently high average accuracy like KS and ASR. However, intermediate tasks (e.g., SID, ER, IC) suffer significant performance degradation, with some methods completely forgetting certain tasks. In contrast, GFL maintains consistently high performance across all tasks, showing strong resistance to forgetting. This is particularly evident in SID, ER, and IC, which are inherently more challenging due to speaker variability and the complexity of intent classification.
% Most methods perform well on tasks with generally high average performance, such as KS and ASR. However, the performance of intermediate tasks significantly degrades, and model even completely forgets certain tasks on some methods. Our method maintains a consistently high performance across all tasks, and also exhibits strong resistance to forgetting. This is particularly evident in SID, ER, and IC. These tasks are often more challenging due to the variability in speaker characteristics,\textcolor{blue}{and?} the complexity of intent classification. 
% 
These results demonstrate that our method balances performance across all CSL tasks while mitigating catastrophic forgetting. The observed positive transfer properties suggest broader multi-task potential, indicating that GFL is a promising component for improving model adaptability and robustness in continual learning.

\subsection{Analysis of Few-shot Pre-training}

\begin{figure}[t]
    \centering
    \includegraphics[width=0.75\linewidth]{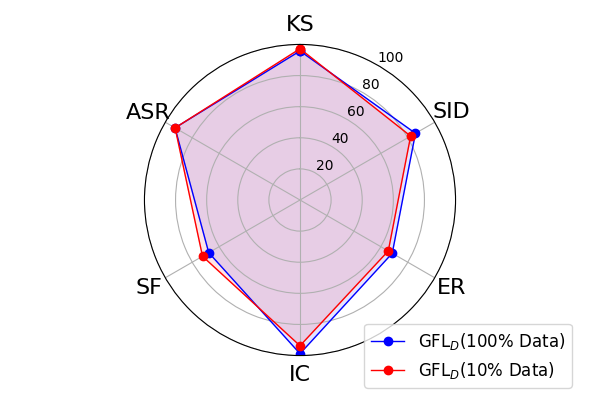}
    \caption{Performance of GFL$_{D}$ initialized by weights pre-trained from different scale of data. 
    }
    \label{fig:few-shot}
\end{figure}

As shown in Figure~\ref{fig:few-shot}, We present a radar chart analysis comparing GFL$_{D}$'s performance under different pre-training data scales, where ASR performance is computed as 100\% minus WER.
GFL$_{D}$ pre-trained under 10\% data maintains nearly 95\% relative performance across all tasks, demonstrating GFL$_{D}$'s exceptional few-shot adaptability.

Notably for task SF, few-shot pre-training achieves 72.36\% STF1 compared to 67.74\% with full-data training, suggesting that the few-shot paradigm may inherently mitigate overfitting by preventing excessive adaptation to the training set.
The proposed few-shot pre-training strategy incurs minimal additional computational overhead compared to standard GFL$_{S}$ training, while providing intrinsic regularization benefits. 
This demonstrates that parameter-efficient adaptation can be achieved without substantial resource investment.

\subsection{Analysis of Prompts}
\label{sec:Prompts}

% Since Whisper is pre-trained for transcription and translation tasks, which excel at generating continuous natural language text, embedding classification labels within a natural language context allows the model to handle these labels more effectively\hao{not clear, explain model is pre-trained on generative tasks, so you convert classification tasks to the form of generative task, and then explain how you do this(add prompts)}. As a result, we attempt to incorporate natural language prompts before classification labels to better leverage the model’s pre-training in natural language generation. 

% Whisper is pre-trained on generative tasks \textcolor{blue}{ASR and ST?} \textcolor{blue}{remove:, which focuses on generating sequences} in the form of natural language. To adapt the model for classification tasks, we decode these tasks in the way of generative tasks. 
Whisper is pre-trained on generative tasks such as ASR and ST, which involve producing natural language sequences. To adapt it for classification tasks, we reformulate them as generative tasks during decoding.
We achieve this by making models generate natural language prompts before predicting labels.
For example, instead of predicting a label like ``happy" directly, the model is required to generate a sentence like ``Emotion is happy.", allowing it to utilize its generative capabilities more effectively.
Our experiments show that prompts can significantly affect model performance across various tasks, as shown in Table~\ref{tab:prompt_decoding_results}.\footnote{SF and ASR are excluded because ASR is a pre-trained task and SF is highly similar to ASR.} 
For complex tasks where the output deviates considerably from the original audio transcribe, such as IC, prompts help guide the model toward more accurate and relevant outputs, making up the gap between pre-training and classification tasks. However, for KS, which closely resembles ASR, direct output without prompts tends to be more effective.
% 
% Further analysis suggests that prompts not only help align the model's semantic understanding in generative tasks, but also make up the gap between pre-training and classification tasks.
% 
% For complex tasks, such as IC, longer prompts provide more contextual understanding \textcolor{blue}{what is longer prompts?}, while for simple and direct tasks, shorter or even no prompts result in more efficient performance. 
% 
Therefore, prompt design must be carefully tailored for each task to achieve the best performance.

% Our experiments demonstrate that the use of prompts can significantly influence the performance of the model across different tasks.Specifically, for tasks where the output significantly deviates from the original audio text, prompts provide substantial improvements. This finding indicates that prompts can help guide the model in generating more accurate and relevant outputs when dealing with tasks that require more complex semantic processing. 

% However, for task KS, which is closely related to ASR, direct output without prompts tends to yield better results.  that for ASR-like tasks, the model benefits from relying directly on the raw audio data without additional contextual guidance. 

% In tasks such as KS, which closely resemble ASR in their output structure, the model benefits from direct reliance on raw audio data without additional contextual guidance, as prompts may introduce unnecessary complexity in these cases.

% Additionally, The choice of whether to include specific words like "The" in prompts also appears to be task-dependent, so it is need for careful prompt design in optimizing model performance.
\begin{table}[t]
    \caption{Impact of prompt formulation and decoding constraints on speech task performance (accuracy). }
    \vspace{-.1cm}
    \centering
    \resizebox{.45 \textwidth}{!}{
        \begin{tabular}{c|c|c|c|c|c}
            \toprule
            Prompt & Decoding & KS & SID & ER & IC \\
            \midrule
            
            None &  & 97.11 & 33.93 & 37.51 & 75.67 \\ 
            ``xxx is ...'' & None & 89.78 & 55.36 & 47.00 & 99.31 \\ 
            ``The xxx is ...'' &  & 89.61 & 51.79 & 56.59 & 95.94 \\ 

            \midrule

            None &  & 97.40 & 39.29 & 39.54 & 76.30 \\ 
            ``xxx is ...'' & Constraint & 89.78 & 58.93 & 47.00 & 99.34 \\ 
            ``The xxx is ...'' &  & 89.61 & 55.36 & 56.41 & 96.02 \\ 
            
            \bottomrule
        \end{tabular}
    }
    \vspace{-.1cm}
    \label{tab:prompt_decoding_results}
\end{table}

\subsection{Analysis of Constrained Decoding}

At inference, we constrain the output space of the classification tasks to the given prompts and possible labels representing the class names.
% The output space of the classification tasks is not the entire vocabulary but consist only of given prompts and possible labels representing the class names. Therefore, for these tasks, we provide prompts and apply constrained decoding during inference to validate its effectiveness.
In Table \ref{tab:prompt_decoding_results}, we observe that constrained decoding significantly improves performance across all the tasks. Restricting the output space to valid labels allows the model to focus on the correct set of possibilities, which enhances accuracy and consistency. In contrast, the model without constrained decoding may generate irrelevant or invalid outputs, especially in complex tasks. By narrowing the output space, we effectively reduce the risk of these false predictions.
This strategy is particularly effective in continual learning setting, where the model is required to handle multiple tasks over time. With constrained decoding, we ensure that the model remains focused on the specific task at hand, reducing the chance of catastrophic forgetting and maintaining higher overall task performance.

\subsection{Analysis of Task Order}

\begin{table}[t]
    \caption{Performance (accuracy) comparison of GFL$_{S}$ and GFL$_{D}$ across different task sequences in continual learning.}
    \vspace{-.1cm}
    \centering
    \resizebox{.45 \textwidth}{!}{
        \begin{tabular}{c|ccc|ccc}
            \toprule
            
            \multirow{2}{*}{Task order} & \multicolumn{3}{c|}{GFL$_{S}$} & \multicolumn{3}{c}{GFL$_{D}$}  \\
            & KS & SID & ER & KS & SID & ER \\
            \midrule
            
            KS - SID - ER & 95.85 & 71.43 & 66.54 & 92.86 & 80.36 & 67.28 \\ 
            KS - ER - SID & 91.01 & 58.93 & 61.47 & 88.87 & 69.64 & 69.49 \\ 
            SID - KS - ER & 88.54 & 82.14 & 66.64 & 89.32 & 83.93 & 63.87 \\ 
            SID - ER - KS & 80.98 & 76.79 & 67.37 & 98.55 & 76.79 & 67.65 \\
            ER - KS - SID & 89.32 & 82.14 & 53.64 & 89.19 & 64.79 & 70.34 \\
            ER - SID - KS & 74.20 & 78.57 & 64.88 & 88.19 & 76.79 & 71.34 \\
            
            \midrule
            MEAN $\uparrow$ & 86.69 & 75.00 & 63.42 & 89.66 & 75.30 & 68.30 \\
            STDEV$\downarrow$ & 7.82 & 8.82 & 5.24 & 1.64 & 7.18 & 2.65 \\
            
            \bottomrule
        \end{tabular}
    }
    \vspace{-.1cm}
    \label{tab:task_order_results}
\end{table}

We study the impact of our method under different orders of task. We choose KS, SID and ER tasks because they are all classification tasks, which focus on different aspects of speech, i.e., content, speaker and emotion, respectively.
Our method demonstrates robust performance across various task sequences in continual learning scenarios, particularly when handling dynamic permutations of KS, SID and ER. Table \ref{tab:task_order_results} presents the results of GSL$_{S}$ and GSL$_{D}$ under different task orders. We also report the mean performance (MEAN) and its standard deviation (STDEV) to evaluate the overall stability.

Both GSL$_{S}$ GSL$_{D}$ achieve high performance, but GSL$_{D}$ maintains stable and superior results regardless of task execution order, validating its resilience to sequential dependencies, enabling reliable deployment in open-world environments with unpredictable task scheduling. 
% \mich{please add a baseline () to the table and explain how GSL$_{D}$ is better than the baseline. I remember you have results for GSL$_{S}$?}

\subsection{Progression of Continual Learning Process}
\begin{figure}[t]
    \centering
    \includegraphics[width=0.95\linewidth]{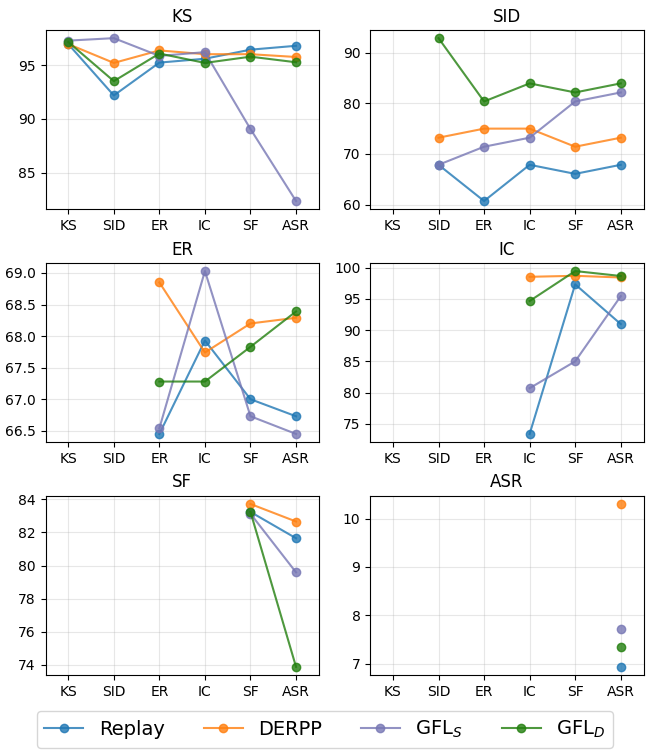}
    \caption{Performance progression of continual learning methods across six tasks. Each curve represents the accuracy (for classification tasks), slot type F1 score (STF1, for SF), or word error rate (WER, for ASR) at different stages of training. 
    }
    \vspace{-.3cm}
    \label{fig:detailed learning process}
\end{figure}

Figure \ref{fig:detailed learning process} compares Replay, DERPP, GFL$_{S}$, and GFL$_{D}$ in detailed continual learning process on the six tasks. 
GFL$_{D}$ exhibits superior task stability with no significant fluctuation. 
The stability originates from its predefined weight freezing mechanism that effectively mitigates parameter drift. 
% Besides, the method demonstrates exceptional performance and improvement. 
Besides, GFL$_{D}$ achieves peak performance all the time on SID, outperforming DERPP and Replay. 
GFL$_{D}$ also shows progressive improvement with regard to ER task, indicating positive knowledge accumulation, which also appears in GFL$_{S}$ on SID. 
The pre-trained weights act as stable anchors that align feature spaces across tasks, enabling effective knowledge inheritance rather than destructive overwriting.
These advantages coalesce into comprehensive performance balance. 
GFL$_{D}$ achieves the high average accuracy with low inter-task variance. 
GFL$_{D}$ is an optimal strategy for speech continual learning deployments.

\section{Conclusion}

We introduced a novel framework for speech continual learning by integrating a Gated-Fusion Layer (GFL) into Whisper, enabling dynamic fusion of speech features for improved adaptability. Our method significantly improves performance across multiple tasks, particularly in knowledge retention and mitigating catastrophic forgetting.
Our results indicate that task-specific prompts and constrained decoding improve certain tasks. In our future work, we will explore advanced fusion mechanisms to further enhance adaptability in evolving speech environments.

% In this work, we introduced a novel framework for speech continual learning in speech processing. By leveraging the Whisper model and incorporating a GFL, we effectively fused speech features from different layers of the encoder, enabling the model to handle diverse speech tasks with improved adaptability. Our method demonstrated significant improvements across multiple tasks, particularly in retaining learned knowledge and mitigating catastrophic forgetting in continual learning scenarios.

% Our findings suggest that the use of task-specific prompts and constrained decoding contributes to better performance in certain tasks, while others benefit more from raw audio inputs. Future work can focus on optimizing prompt design and exploring more advanced fusion mechanisms to further enhance the model's performance in dynamic and evolving speech environments.

% An interesting direction for future exploration is investigating how knowledge acquired from one type of task (e.g., speaker identification) can be transferred to improve performance on other related tasks (e.g., emotion recognition or intent classification). Techniques such as transfer learning or meta-learning could be explored to enhance the model's ability to generalize knowledge across tasks, leading to more efficient learning in scenarios with limited data for new tasks.

\section{Acknowledgements}
% Acknowledgement should only be included in the camera-ready version, not in the version submitted for review. The 5th page is reserved exclusively for acknowledgements and  references. No other content must appear on the 5th page. Appendices, if any, must be within the first 4 pages. The acknowledgments and references may start on an earlier page, if there is space.
This work is partially supported by National Nature Science Foundation of China under No. U21A20488. We thank the Big Data Computing Center of Southeast University for providing the facility support on the numerical calculations in this paper. 

The initial draft of this paper was written manually. ChatGPT was later used to enhance clarity and readability. All content was carefully reviewed and verified.

% \ifinterspeechfinal
%      The Interspeech 2025 organisers
% \else
%      The authors
% \fi
% would like to thank ISCA and the organising committees of past Interspeech conferences for their help and for kindly providing the previous version of this template.

\bibliographystyle{IEEEtran}
\bibliography{mybib}

% Generated by IEEEtran.bst, version: 1.13 (2008/09/30)
\begin{thebibliography}{10}
\providecommand{\url}[1]{#1}
\csname url@samestyle\endcsname
\providecommand{\newblock}{\relax}
\providecommand{\bibinfo}[2]{#2}
\providecommand{\BIBentrySTDinterwordspacing}{\spaceskip=0pt\relax}
\providecommand{\BIBentryALTinterwordstretchfactor}{4}
\providecommand{\BIBentryALTinterwordspacing}{\spaceskip=\fontdimen2\font plus
\BIBentryALTinterwordstretchfactor\fontdimen3\font minus \fontdimen4\font\relax}
\providecommand{\BIBforeignlanguage}[2]{{%
\expandafter\ifx\csname l@#1\endcsname\relax
\typeout{** WARNING: IEEEtran.bst: No hyphenation pattern has been}%
\typeout{** loaded for the language `#1'. Using the pattern for}%
\typeout{** the default language instead.}%
\else
\language=\csname l@#1\endcsname
\fi
#2}}
\providecommand{\BIBdecl}{\relax}
\BIBdecl

\bibitem{abs-2402-01364}
T.~Wu, L.~Luo, Y.~Li, S.~Pan, T.~Vu, and G.~Haffari, ``Continual learning for large language models: {A} survey,'' \emph{CoRR}, vol. abs/2402.01364, 2024.

\bibitem{SustekSH22}
\BIBentryALTinterwordspacing
M.~Sustek, S.~Sadhu, and H.~Hermansky, ``Dealing with unknowns in continual learning for end-to-end automatic speech recognition,'' in \emph{23rd Annual Conference of the International Speech Communication Association, Interspeech 2022, Incheon, Korea, September 18-22, 2022}, H.~Ko and J.~H.~L. Hansen, Eds.\hskip 1em plus 0.5em minus 0.4em\relax {ISCA}, 2022, pp. 1046--1050. [Online]. Available: \url{https://doi.org/10.21437/Interspeech.2022-11139}
\BIBentrySTDinterwordspacing

\bibitem{WangZSZ24}
\BIBentryALTinterwordspacing
L.~Wang, X.~Zhang, H.~Su, and J.~Zhu, ``A comprehensive survey of continual learning: Theory, method and application,'' \emph{{IEEE} Trans. Pattern Anal. Mach. Intell.}, vol.~46, no.~8, pp. 5362--5383, 2024. [Online]. Available: \url{https://doi.org/10.1109/TPAMI.2024.3367329}
\BIBentrySTDinterwordspacing

\bibitem{ke2022continual}
Z.~Ke and B.~Liu, ``Continual learning of natural language processing tasks: A survey,'' \emph{arXiv preprint arXiv:2211.12701}, 2022.

\bibitem{qu2021recent}
H.~Qu, H.~Rahmani, L.~Xu, B.~Williams, and J.~Liu, ``Recent advances of continual learning in computer vision: An overview,'' \emph{arXiv preprint arXiv:2109.11369}, 2021.

\bibitem{abs-2401-16386}
D.~Zhou, H.~Sun, J.~Ning, H.~Ye, and D.~Zhan, ``Continual learning with pre-trained models: {A} survey,'' \emph{CoRR}, vol. abs/2401.16386, 2024.

\bibitem{abs-2302-00487}
L.~Wang, X.~Zhang, H.~Su, and J.~Zhu, ``A comprehensive survey of continual learning: Theory, method and application,'' \emph{CoRR}, vol. abs/2302.00487, 2023.

\bibitem{rolnick2019experience}
D.~Rolnick, A.~Ahuja, J.~Schwarz, T.~Lillicrap, and G.~Wayne, ``Experience replay for continual learning,'' \emph{Advances in Neural Information Processing Systems}, vol.~32, 2019.

\bibitem{lopez2017gradient}
D.~Lopez-Paz and M.~Ranzato, ``Gradient episodic memory for continual learning,'' \emph{Advances in neural information processing systems}, vol.~30, 2017.

\bibitem{chaudhry2018efficient}
\BIBentryALTinterwordspacing
A.~Chaudhry, M.~Ranzato, M.~Rohrbach, and M.~Elhoseiny, ``Efficient lifelong learning with {A-GEM},'' in \emph{7th International Conference on Learning Representations, {ICLR} 2019, New Orleans, LA, USA, May 6-9, 2019}.\hskip 1em plus 0.5em minus 0.4em\relax OpenReview.net, 2019. [Online]. Available: \url{https://openreview.net/forum?id=Hkf2\_sC5FX}
\BIBentrySTDinterwordspacing

\bibitem{kirkpatrick2017overcoming}
J.~Kirkpatrick, R.~Pascanu, N.~Rabinowitz, J.~Veness, G.~Desjardins, A.~A. Rusu, K.~Milan, J.~Quan, T.~Ramalho, A.~Grabska-Barwinska \emph{et~al.}, ``Overcoming catastrophic forgetting in neural networks,'' \emph{Proceedings of the national academy of sciences}, vol. 114, no.~13, pp. 3521--3526, 2017.

\bibitem{li2017learning}
Z.~Li and D.~Hoiem, ``Learning without forgetting,'' \emph{IEEE transactions on pattern analysis and machine intelligence}, vol.~40, no.~12, pp. 2935--2947, 2017.

\bibitem{mallya2018piggyback}
A.~Mallya, D.~Davis, and S.~Lazebnik, ``Piggyback: Adapting a single network to multiple tasks by learning to mask weights,'' in \emph{Proceedings of the European conference on computer vision (ECCV)}, 2018, pp. 67--82.

\bibitem{serra2018overcoming}
J.~Serra, D.~Suris, M.~Miron, and A.~Karatzoglou, ``Overcoming catastrophic forgetting with hard attention to the task,'' in \emph{International conference on machine learning}.\hskip 1em plus 0.5em minus 0.4em\relax PMLR, 2018, pp. 4548--4557.

\bibitem{wang2022learning}
Z.~Wang, Z.~Zhang, C.-Y. Lee, H.~Zhang, R.~Sun, X.~Ren, G.~Su, V.~Perot, J.~Dy, and T.~Pfister, ``Learning to prompt for continual learning,'' in \emph{Proceedings of the IEEE/CVF Conference on Computer Vision and Pattern Recognition}, 2022, pp. 139--149.

\bibitem{radford2023robust}
A.~Radford, J.~W. Kim, T.~Xu, G.~Brockman, C.~McLeavey, and I.~Sutskever, ``Robust speech recognition via large-scale weak supervision,'' in \emph{International conference on machine learning}.\hskip 1em plus 0.5em minus 0.4em\relax PMLR, 2023, pp. 28\,492--28\,518.

\bibitem{yang2023investigating}
H.~Yang, J.~Zhao, G.~Haffari, and E.~Shareghi, ``Investigating pre-trained audio encoders in the low-resource condition,'' \emph{arXiv preprint arXiv:2305.17733}, 2023.

\bibitem{WuCLLQH22}
\BIBentryALTinterwordspacing
T.~Wu, M.~Caccia, Z.~Li, Y.~Li, G.~Qi, and G.~Haffari, ``Pretrained language model in continual learning: {A} comparative study,'' in \emph{The Tenth International Conference on Learning Representations, {ICLR} 2022, Virtual Event, April 25-29, 2022}.\hskip 1em plus 0.5em minus 0.4em\relax OpenReview.net, 2022. [Online]. Available: \url{https://openreview.net/forum?id=figzpGMrdD}
\BIBentrySTDinterwordspacing

\bibitem{wang2024comprehensive}
L.~Wang, X.~Zhang, H.~Su, and J.~Zhu, ``A comprehensive survey of continual learning: theory, method and application,'' \emph{IEEE Transactions on Pattern Analysis and Machine Intelligence}, 2024.

\bibitem{yang2021superb}
S.-w. Yang, P.-H. Chi, Y.-S. Chuang, C.-I.~J. Lai, K.~Lakhotia, Y.~Y. Lin, A.~T. Liu, J.~Shi, X.~Chang, G.-T. Lin \emph{et~al.}, ``Superb: Speech processing universal performance benchmark,'' \emph{arXiv preprint arXiv:2105.01051}, 2021.

\bibitem{WuWZLQLH22}
\BIBentryALTinterwordspacing
T.~Wu, G.~Wang, J.~Zhao, Z.~Liu, G.~Qi, Y.~Li, and G.~Haffari, ``Towards relation extraction from speech,'' in \emph{Proceedings of the 2022 Conference on Empirical Methods in Natural Language Processing, {EMNLP} 2022, Abu Dhabi, United Arab Emirates, December 7-11, 2022}, 2022, pp. 10\,751--10\,762. [Online]. Available: \url{https://doi.org/10.18653/v1/2022.emnlp-main.738}
\BIBentrySTDinterwordspacing

\bibitem{kang2024event}
\BIBentryALTinterwordspacing
J.~Kang, T.~Wu, J.~Zhao, G.~Wang, G.~Qi, Y.~Li, and G.~Haffari, ``Towards event extraction from speech with contextual clues,'' \emph{CoRR}, vol. abs/2401.15385, 2024. [Online]. Available: \url{https://doi.org/10.48550/arXiv.2401.15385}
\BIBentrySTDinterwordspacing

\bibitem{warden2017speech}
P.~Warden, ``Speech commands: A public dataset for single-word speech recognition,'' \emph{Dataset available from http://download. tensorflow. org/data/speech\_commands\_v0}, vol.~1, 2017.

\bibitem{nagrani2020voxceleb}
A.~Nagrani, J.~S. Chung, W.~Xie, and A.~Zisserman, ``Voxceleb: Large-scale speaker verification in the wild,'' \emph{Computer Speech \& Language}, vol.~60, p. 101027, 2020.

\bibitem{busso2008iemocap}
C.~Busso, M.~Bulut, C.-C. Lee, A.~Kazemzadeh, E.~Mower, S.~Kim, J.~N. Chang, S.~Lee, and S.~S. Narayanan, ``Iemocap: Interactive emotional dyadic motion capture database,'' \emph{Language resources and evaluation}, vol.~42, pp. 335--359, 2008.

\bibitem{lugosch2019speech}
L.~Lugosch, M.~Ravanelli, P.~Ignoto, V.~S. Tomar, and Y.~Bengio, ``Speech model pre-training for end-to-end spoken language understanding,'' \emph{arXiv preprint arXiv:1904.03670}, 2019.

\bibitem{lai2021semi}
C.-I. Lai, Y.-S. Chuang, H.-Y. Lee, S.-W. Li, and J.~Glass, ``Semi-supervised spoken language understanding via self-supervised speech and language model pretraining,'' in \emph{ICASSP 2021-2021 IEEE International Conference on Acoustics, Speech and Signal Processing (ICASSP)}.\hskip 1em plus 0.5em minus 0.4em\relax IEEE, 2021, pp. 7468--7472.

\bibitem{panayotov2015librispeech}
V.~Panayotov, G.~Chen, D.~Povey, and S.~Khudanpur, ``Librispeech: an asr corpus based on public domain audio books,'' in \emph{2015 IEEE international conference on acoustics, speech and signal processing (ICASSP)}.\hskip 1em plus 0.5em minus 0.4em\relax IEEE, 2015, pp. 5206--5210.

\bibitem{buzzega2020dark}
P.~Buzzega, M.~Boschini, A.~Porrello, D.~Abati, and S.~Calderara, ``Dark experience for general continual learning: a strong, simple baseline,'' \emph{Advances in neural information processing systems}, vol.~33, pp. 15\,920--15\,930, 2020.

\bibitem{chaudhry2018riemannian}
A.~Chaudhry, P.~K. Dokania, T.~Ajanthan, and P.~H. Torr, ``Riemannian walk for incremental learning: Understanding forgetting and intransigence,'' in \emph{Proceedings of the European conference on computer vision (ECCV)}, 2018, pp. 532--547.

\bibitem{loshchilov2017decoupled}
I.~Loshchilov and F.~Hutter, ``Decoupled weight decay regularization,'' \emph{arXiv preprint arXiv:1711.05101}, 2017.

\end{thebibliography}

\end{document}